\PassOptionsToPackage{table}{xcolor} 
\documentclass[a4paper,fleqn]{cas-dc}
\usepackage{graphicx} 
\usepackage[utf8]{inputenc}
\usepackage[T1]{fontenc}
\usepackage{orcidlink}
\usepackage[authoryear,longnamesfirst]{natbib}
\usepackage{float}
\usepackage{caption} 
\usepackage{algorithm}
\usepackage{algpseudocode} 
\usepackage{hyperref}
\makeatletter
\AtBeginDocument{\expandafter\let\csname__first_footerline:\endcsname\relax}
\makeatother

\begin{document}
\let\WriteBookmarks\relax
\def\floatpagepagefraction{1}
\def\textpagefraction{.001}

\shorttitle{Data-Efficient Deep Learning: Empirical Guidelines for Training Set Size Estimation in Inertial Sensor Classification}
\shortauthors{O. Kruzel and I. Klein}

\title[mode = title]{Data-Efficient Deep Learning: Empirical Guidelines for Training Set Size Estimation in Inertial Sensor Classification}

\author[1]{Ofir Kruzel}[orcid=0009-0000-4015-0853]
\cormark[1] 
\ead{okruzeld@campus.haifa.ac.il}

\author[1]{Itzik Klein}[orcid=0000-0001-7846-0654]

\affiliation[1]{organization={The Autonomous Navigation and Sensor Fusion Lab, the Hatter Department of Marine Technologies, University of Haifa},
                addressline={}, 
                city={Haifa},
                country={Israel}}

\cortext[1]{Corresponding author}

\begin{abstract}
Deep learning models’ dependency on large-scale inertial datasets presents a significant bottleneck in inertial sensor-based classification tasks, such as human activity recognition and smartphone location recognition. In these domains, data collection requires massive recording campaigns that are complex, time-consuming, and difficult to scale. Currently, data-driven guidelines for determining the minimum sample size required to reach a desired accuracy level do not exist. To address this gap, this study presents a systematic empirical evaluation of learning curve convergence rates in inertial classification. We introduce a unified framework that analyzes classification performance under both binary and multi-class scenarios, and derive an empirical formula to estimate performance relative to dataset size. Testing across six diverse, real-world datasets totaling 102.7 hours of inertial measurements demonstrates that accuracy follows a consistent logarithmic growth pattern, regardless of task complexity. Leveraging this finding, we propose a quantitative stability point metric, defined as the sample size required for the learning curve to stabilize within a predefined mean absolute percentage deviation of its asymptotic maximum. Our analysis reveals that models often reach practical stability with substantially fewer samples than traditional heuristics suggest. Ultimately, we offer a generalizable framework to extrapolate total data requirements from small-scale pilot studies, optimizing the tradeoff between recording effort and model reliability. These findings shift the prevailing paradigm from maximizing data volume toward optimizing data efficiency, offering concrete, data-backed guidelines for planning recording campaigns in inertial sensing applications.

\end{abstract}

\begin{highlights}
\item Evaluates learning curve convergence for inertial sensor classification.
\item Analyzes six diverse HAR and SLR datasets totaling 102.7 hours of data. 
\item Reveals classification accuracy consistently follows a logarithmic pattern.
\item Introduces a stability point metric to optimize training data collection. 
\item Shows models reach stability with fewer samples than heuristics suggest.
\end{highlights}



\maketitle

\section{Introduction}
Human recognition from wearable and smartphone sensors has become a core capability in ubiquitous and mobile computing. By exploiting accelerometer and gyroscope measurements, these systems infer user activities, motion modes, and device carrying configurations \cite{cuesta2010use, camomilla2018trends}, supporting applications in healthcare, sports analytics, human–computer interaction, and context aware navigation. \cite{klein2025pedestrian, yampolsky2025neural, MARDANPOUR2023119073}. Over the past decade, deep learning models, such as convolution, recurrent, and attention based architectures, have substantially improved recognition accuracy and robustness, leading to a rich body of work on model architectures, sensor configurations, and deployment scenarios \cite{gu2021survey,kaseris2024survey,gomaa2023perspective, vertzberger2021attitude}.


Despite these advances, comparatively little attention has been paid to a basic practical question shared across activity and location recognition tasks: how much labeled sensor data is actually required to reach a desired level of classification performance? The relationship between training dataset size and classification accuracy remains a fundamental yet practically challenging issue in machine learning. While it is well understood that model performance generally improves with more data \cite{cortes1993learning}, a more granular question arises: how much data is “enough” to achieve reliable accuracy without incurring excessive data collection? This question is especially critical in domains such as human activity recognition (HAR) \cite{make6020040, cohen2024inertial} and smartphone location recognition (SLR) \cite{klein2019smartphone}; where collecting and labeling sensor data is labor intensive, often requiring precise temporal segmentation and video based ground truth. Consequently, the engineering objective shifts from maximizing data volume to optimizing data efficiency determining the minimum data size sufficient to achieve robust performance \cite{melvin2021sample,fekson2026enhancement,kruzel2026optimizing}. This bottleneck is not unique to inertial sensing. For instance, in agricultural deep learning, researchers have noted that defining exactly how much data is enough remains a major challenge due to the immense effort required to capture and accurately label real-world samples \cite{barbedo2018impact, abbas2026bibliometric}.

A longstanding guideline is the ``Rule of 10'', which suggests that a dataset should contain at least ten samples for every feature or parameter in the model \cite{abumostafa1989information, raudys1991small}. While useful for simple linear regressions, this heuristic is generally insufficient for modern high dimensional deep learning models, where simplistic sample size guidelines have been shown to be inadequate in both clinical prediction \cite{dhiman2023sample} and machine learning contexts \cite{balki2019sample}. In fact, systematic evaluations in the medical domain confirm that despite continuous efforts, there is still no universally agreed-upon definition of what even constitutes a small dataset. Crucially, these evaluations reveal that the overall performance of a classifier actually depends more on the extent to which a dataset represents the original data distribution rather than its sheer size \cite{althnian2021impact}. Consequently, researchers have turned to learning curve analysis to model the trajectory of performance improvement. Seminal theoretical work established that error rates typically follow a Power Law decline as training set size increases \cite{cortes1993learning, hutter2021learning}. Recent advancements have refined these into formal ``Scaling Laws''. Notably, the work of Rosenfeld et al. \cite{rosenfeld2019constructive} established a constructive functional form to predict generalization error as a simultaneous function of both dataset size ($n$) and the number of model parameters ($m$).

However, these laws are often optimized for large scale vision or language tasks. In domain specific engineering applications, fixed architectures and require precise, data driven thresholds rather than broad asymptotic bounds \cite{rajput2023evaluation}. In practical engineering contexts, particularly those involving bounded metrics like classification accuracy, the classical power law assumption often fails to capture the saturation and “diminishing returns” phases observed in smaller, noisy datasets. Systematic analyses of learning curves have shown that they can be non monotonic or otherwise irregular, underscoring that simple power law formulations are not universally valid \cite{loog2022survey}. Recent empirical work in tabular clinical classification further demonstrates that accuracy growth is frequently better described by a logarithmic model, which more naturally reflects a stability point beyond which additional data yields negligible gains \cite{silvey2024sample}.

Despite these insights, a critical gap remains in the HAR and SLR domain. Factors like feature dimensionality, and activity class granularity (binary vs. multi class) significantly influence the convergence rate, yet there is no unified, generalizable model to predict this trajectory across diverse real world datasets. In this work, we address this gap by conducting a systematic empirical study of how training dataset size impacts classification performance across six different datasets spanning HAR and SLR tasks. Our methodology is designed to enable comparison of convergence behavior despite the disparate natures of the datasets. For each dataset, we evaluate classification accuracy under two scenario variants (binary classification versus multi class classification) and measure performance at ten incrementally increasing sample sizes on a uniform measuring scale. This unified evaluation framework allows us to observe and quantify the convergence pattern of accuracy in each case.

We present a unified evaluation framework that analyzes learning curve convergence across diverse real world inertial datasets, enabling researchers to extrapolate total data requirements from small scale pilot studies and reduce the burden of massive recording campaigns that are complex, expressive, and time consuming. The key contribution of this work is the empirical identification of a consistent logarithmic growth pattern for bounded inertial sensor classification tasks. building on this, we offer a data-driven framework enabling researchers to optimizing the trade off between recording effort and model reliability.
 
Our main results formally characterize this behavior across all evaluated datasets without imposing any specific functional form \textit{a priori}, we find that the data from every case is well described by a logarithmic function of the form: $\mathrm{Accuracy}(n)=a\log(n)+b$ where $n$ is the sample size and $a, b$ are constants. This suggests a common ``logarithmic regime'' governing performance gains in sensor based tasks.

Establishing a quantitative stability point directly addresses the resource intensive nature of data acquisition in inertial sensing. This metric provides vital technical insights into the parameters governing model convergence, allowing for a systematic evaluation of when additional data collection yields diminishing returns. From an engineering perspective, this approach is critical for reducing the significant temporal and financial investments required for manual annotation and sensor recording. By identifying the minimum sufficient sample size needed to satisfy performance requirements, this framework enables a more efficient allocation of laboratory resources and prevents redundant data collection during the development of classification systems.

The remainder of this paper is organized as follows: Section \ref{sec:datasets} presents the characteristics of the six datasets and the unified experimental framework, including the deep learning architectures and training protocols. Section  \ref{sec:tasks} details the research approach proposed in this study. Section \ref{sec:results} presents the empirical learning curves for both binary and multi-class scenarios, analyzing the fit of the logarithmic model. Finally, Section \ref{sec:conclusions} gives the conclusions of this research.

\section{Datasets} \label{sec:datasets}

We conducted our study across six publicly available datasets spanning HAR and SLR tasks. The technical specifications, including subject counts, sampling rates, and total durations, are summarized in Table \ref{tab:dataset_summary}: 

\begin{enumerate}
    \item \textbf{PAMAP2}: This dataset is utilized to evaluate model convergence in scenarios requiring high activity granularity and multi-location sensor fusion. It features a wide variety of daily and vigorous exercise activities captured via multiple body-worn IMUs \cite{pamap2_physical_activity_monitoring_231}. 

    \item \textbf{MotionSense}: Focuses on smartphone-based sensing within a pocket-worn configuration. It is particularly valuable for this study due to its demographic diversity, providing variability in how activities are performed across different participant profiles \cite{malekzadeh2019mobile}.

    \item \textbf{MobilePos}: Unlike the activity-centric sets, this dataset provides a dedicated focus on SLR. It allows us to test the consistency of our logarithmic growth findings on tasks where phone placement is the primary classification target \cite{yan2018ridi}.

    \item \textbf{REALDISP}: This set is included to analyze the influence of realistic sensor displacement and high activity diversity. With over thirty distinct activities, it represents one of the most complex classification challenges in our experimental framework \cite{uddin2015human}.

    \item \textbf{UCI-HAR}: This dataset serves as a standardized baseline. It enables us to validate our unified framework against traditional inertial sensing results in a highly controlled, smartphone-based environment \cite{anguita2013public}.

    \item \textbf{WISDM}: This dataset is selected for its scale, providing a large participant pool and extensive recording duration. It is critical for observing how learning curves behave as they approach their true asymptotic maximum in a large-scale data mining context \cite{weiss2019wisdm}.
\end{enumerate}

Table~\ref{tab:dataset_summary} provides an overview of key dataset characteristics. In total the datasets contains recording of 142 subjects with a total duration of 102.75 hours.

\begin{center}
\captionof{table}{Summary of Datasets Used in the Study.}
\label{tab:dataset_summary}
\resizebox{\columnwidth}{!}{%
\begin{tabular}{|l|c|c|c|c|}
\hline
\textbf{Dataset} & \textbf{Subjects} & \textbf{Activities} &  \textbf{Duration (hours)} & \textbf{Sampling (Hz)} \\
\hline
PAMAP2 & 9 & 18 &  41.7 & 100 \\
MotionSense & 24 & 6 & 7.25 & 50 \\
MobilePos & 10 & 4 & 2.5 & 200 \\
REALDISP & 17 & 33 & 17 & 100 \\
UCI-HAR & 30 & 6 & 1.8 & 50 \\
WISDM & 52 & 18 & 32.5 & 20 \\
TOTAL & 142 & 103 & 102.75 & 20-200 \\
\hline

\end{tabular}%
}
\end{center}

\section{Proposed Approach} \label{sec:tasks}

The primary objective of this study is to identify a robust mathematical characterization of the relationship between training dataset size and classification in inertial sensing. Our research was motivated by the theoretical foundations established in large-scale vision and language modeling \cite{cortes1993learning, hutter2021learning, rosenfeld2019constructive}. These foundational works typically model generalization error as a complex interaction between both dataset size and the number of model parameters.

The goal of this work was to determine if a simplified version of these estimations could be generalized to inertial sensor classification, specifically for practical engineering contexts where researchers prioritize data volume requirements over extensive architectural tuning. To this end, we conducted a systematic evaluation of several candidate approximation formulas, including power-law, exponential, and polynomial functions. We tested these functions by removing their explicit dependency on model parameters to isolate the influence of training sample size. 

Our preliminary empirical analysis across diverse inertial datasets revealed that once model parameter dependency was removed, traditional power-law, exponential, and polynomial forms failed to accurately capture the performance growth trajectory. In contrast, we found that a logarithmic growth pattern provided the most consistent and stable fit for bounded classification accuracy. 

Based on these findings, we focus our evaluation framework on the following logarithmic model:
\begin{equation}
\label{eq:accuracy_model_final}
\mathrm{Accuracy}(n) = a \log(n) + b
\end{equation}
where $n$ represents the number of training samples, and $a$ and $b$ are constants determined by the specific task scenario, respectively.

In the following sections we elaborate on our methodology, baseline network structure, and training process.

\subsection{Methodology} \label{sec:methodology}

Our methodology follows a systematic empirical study designed to identify and quantify the convergence behavior of classification accuracy as a function of training dataset size. To ensure the findings are generalizable across different problem complexities, we employ a unified evaluation framework that processes diverse real-world inertial datasets through a standardized pipeline. 

\begin{figure*}[!tb]
\centering
\includegraphics[width=\textwidth]{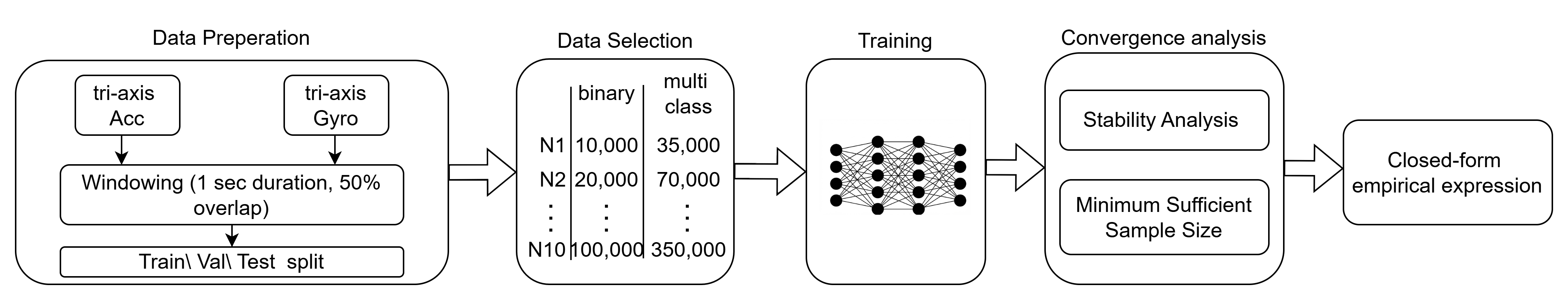}
\caption{Experimental methodology flowchart showing data preparation and selection, training and convergence analysis.}
\label{fig:chart}
\end{figure*}

As illustrated in our experimental framework flowchart (Figure~\ref{fig:chart}), this process involves four primary stages: (i) raw sensor data preparation through sliding window segmentation, (ii) division into parallel binary and multi-class scenarios, (iii) model training across an incrementally scaled grid of training samples, and (iv) curve modeling and stability analysis to identify the practical convergence point. This systematic approach isolates training data volume as the independent variable while keeping architectural and optimization settings constant across all experimental configurations.

\subsection{Network Structure} \label{sec:architecture}

The baseline architecture is a CNN-BiLSTM network that maps a fixed-length window of raw IMU signals to class probabilities through three sequential stages: a convolutional front end for local feature extraction, a BiLSTM encoder for temporal context modeling, and a fully connected classification head. The architecture consists of the following blocks:
 
1. \textbf{Input}:
Each input sample corresponds to a fixed-length window of inertial sensor readings consisting of tri-axial accelerometer and gyroscope signals. Let
\begin{gather}
\mathbf{f}_k = [f_{x,k}\ f_{y,k}\ f_{z,k}]^{T}, \label{eq:imu_vectors1}\\
\boldsymbol{\omega}_k = [\omega_{x,k}\ \omega_{y,k}\ \omega_{z,k}]^{T} \label{eq:imu_vectors2}
\end{gather}
denote the specific force and angular velocity vectors at time $k$, respectively. These are concatenated to form a 6-dimensional observation
\begin{equation}
\label{eq:observation_vector}
\mathbf{x}_k = [\mathbf{f}_k^\top, \boldsymbol{\omega}_k^\top]^\top \in \mathbb{R}^6
\end{equation}
For a window of length $j$ time steps, the network input is the sequence
\begin{equation}
\label{eq:input_window}
\mathbf{X} = \{\mathbf{x}_1, \dots, \mathbf{x}_j\} \in \mathbb{R}^{j \times 6}
\end{equation}
which jointly encodes translational and rotational motion over a short temporal context. This representation is used uniformly across all datasets and both task scenarios (binary and multi-class), so that differences in convergence behavior can be attributed to dataset characteristics and label granularity rather than to changes in model input.
 
2. \textbf{Conventional frontend:}
A 1D convolutional layer slides learned filters across the time axis of $\mathbf{X}$ to extract local temporal features. 

Let $\mathbf{w}_m \in \mathbb{R}^{C \times L}$ be the kernel weights of the $m$-th filter of temporal length $l$, with a corresponding scalar bias $b_m \in \mathbb{R}$. The cross-correlation output of filter $m$ at discrete time step $k$ is expressed as:
\begin{equation}
\label{eq:convolution_1d}
z_{m,k} = (\mathbf{X} * \mathbf{w}_m)_k = \sum_{c=1}^{C} \sum_{p=0}^{l-1} w_{m,c,p} \, x_{c,k+p} + b_m
\end{equation}

where $*$ denotes discrete 1D cross-correlation in the deep learning sense \cite{sze2017efficient}. In our network, $\mathbf{X}$ has $C=6$ channels and is processed by $M=64$ filters with kernel size 5 and stride 1, yielding a feature map $\mathbf{z} \in \mathbb{R}^{64 \times K'}$.

The feature map is then passed through a ReLU nonlinearity \cite{sharma2017activation},
\begin{equation}
\label{eq:relu_activation}
\mathrm{ReLU}(u) = \max(0, u)
\end{equation}
where the mapping is executed elementwise across all feature dimensions. followed by 1D max pooling with window size $d$ and stride $s_p$, which retains the most salient response in each local window \cite{boureau2010theoretical}:
\begin{equation}
\label{eq:max_pooling}
z^{\mathrm{p}}_{m,k} = \max_{0 \le i < d} z_{m,\, s_p k + i}
\end{equation}
With $d=3$ and $s_p=3$, the pooled temporal length reduces to
\begin{equation}
\label{eq:pooled_length}
\tilde{K} = \left\lfloor \frac{L - d}{s_p} \right\rfloor + 1
\end{equation}
where $L$ is the input temporal length. This yields a sequence of pooled feature vectors
\begin{equation}
\label{eq:pooled_feature_vectors}
\mathbf{Z}^{\mathrm{p}} = \{\mathbf{z}^{\mathrm{p}}_1, \dots, \mathbf{z}^{\mathrm{p}}_{\tilde{K}}\}, \quad \mathbf{z}^{\mathrm{p}}_k \in \mathbb{R}^{64}, \quad \mathbf{Z}^{\mathrm{p}} \in \mathbb{R}^{\tilde{K} \times 64}
\end{equation}
 
3. \textbf{BiLSTM encoder:}
To capture temporal dependencies in both directions, the pooled sequence $\mathbf{Z}^{\mathrm{p}}$ is processed by a BiLSTM block. A forward LSTM reads the sequence left to right, computing hidden states
\begin{equation}
\label{eq:lstm_forward}
\overrightarrow{\mathbf{h}}_k = \overrightarrow{\mathrm{LSTM}}\big(\mathbf{z}^{\mathrm{p}}_k,\, \overrightarrow{\mathbf{h}}_{k-1}\big) \in \mathbb{R}^{H}
\end{equation}
while a backward LSTM processes the sequence in reverse,
\begin{equation}
\label{eq:lstm_backward}
\overleftarrow{\mathbf{h}}_k = \overleftarrow{\mathrm{LSTM}}\big(\mathbf{z}^{\mathrm{p}}_k,\, \overleftarrow{\mathbf{h}}_{k+1}\big) \in \mathbb{R}^{H}
\end{equation}
The BiLSTM concatenates both directions at each time step \cite{hochreiter1997long,graves2005framewise},
\begin{equation}
\label{eq:bilstm_concatenation}
\mathbf{h}_k = \mathrm{BiLSTM}\big(\mathbf{Z}^{\mathrm{p}}\big)_k =
\begin{bmatrix}
\overrightarrow{\mathbf{h}}_k \\
\overleftarrow{\mathbf{h}}_k
\end{bmatrix}
\in \mathbb{R}^{2H}
\end{equation}
thus encoding past and future context jointly at each position. With $H=128$, the full output sequence $\mathbf{h}_k \in \mathbb{R}^{256}$ is passed to the classification head such that:
\begin{equation}
\label{eq:bilstm_output_seq}
\mathbf{h}_k = \mathrm{BiLSTM}\big(\mathbf{Z}^{\mathrm{p}}\big)_k \in \mathbb{R}^{256}
\end{equation}
 
4. \textbf{Classification head:}
The BiLSTM output sequence is flattened into a single feature vector and regularized with dropout \cite{srivastava2014dropout}. Given a hidden vector $\mathbf{h} \in \mathbb{R}^{d}$ and dropout rate $p \in [0,1)$, a binary mask $\mathbf{r} \in \{0,1\}^{d}$ with independent components
\begin{equation}
\label{eq:dropout_mask}
r_{j} \sim \mathrm{Bernoulli}(1-p), \quad j=1, \dots, d
\end{equation}
is sampled and applied elementwise,
\begin{equation}
\label{eq:dropout_hadamard}
\tilde{\mathbf{h}} = \mathbf{h} \odot \mathbf{r}
\end{equation}
where $\odot$ denotes the Hadamard product. With $p=0.25$, let $\tilde{\mathbf{h}}^{\mathrm{flat}}$ denote the resulting regularized feature vector. A fully connected layer maps this to class logits,
\begin{equation}
\label{eq:final_logits}
\mathbf{y} = \mathbf{W}_{\mathrm{FC}} \tilde{\mathbf{h}}^{\mathrm{flat}} + \mathbf{b}_{\mathrm{FC}}
\end{equation}
where $\mathbf{W}_{\mathrm{FC}} \in \mathbb{R}^{C_{\text{out}} \times 256}$, $\mathbf{b}_{\mathrm{FC}} \in \mathbb{R}^{C_{\text{out}}}$, and $C_{\text{out}}$ is the number of classes (two for binary tasks, or the full class count for multi-class tasks). Finally, a softmax layer \cite{bridle1990probabilistic} converts the logits to class probabilities,
\begin{equation}
\label{eq:softmax_output}
\mathrm{softmax}(\mathbf{y})_c = \frac{\exp(y_c)}{\sum_{j=1}^{C_{\text{out}}} \exp(y_j)}, \quad c = 1,\dots,C_{\text{out}}
\end{equation}
\subsection{Training} \label{sec:design}

Our experiment design was constructed to (i) place all datasets on a comparable measurement scale, and (ii) enable a controlled analysis of how classification performance evolves as a function of the number of training samples. To this end, we first transformed each continuous sensor stream into fixed length windows, and then derived ten incremental training set sizes for both the binary and multi class scenarios.
Algorithm~\ref{alg:training} summarizes the full procedure, and the subsections below detail each step.

\begin{algorithm}[t]
\caption{Experimental Training Procedure}
\label{alg:training}
\begin{algorithmic}[1]
\renewcommand{\algorithmicrequire}{\textbf{Input:}}
\renewcommand{\algorithmicensure}{\textbf{Output:}}
\Require $\mathbb{D}=\{\mathcal{D}_i\}$ with rates $f_{s,i}$\,[Hz];\quad
         $\mathcal{S}=\{\mathrm{bin},\mathrm{mc}\}$;\quad
         grids $\mathcal{N}^{s}=\{n_1,\dots,n_{10}\}$;\quad runs $R$
\Ensure  $\big\{(n_i,\,\overline{\mathrm{Accuracy}}(n_i))\big\}_{i=1}^{10}$ per $(\mathcal{D},s)$
\Statex
\Statex \Comment{\textit{Data Preparation} — Sec.~\ref{sec:windowconstruction}, Eq.~\eqref{eq:input_window}}
\ForAll{$\mathcal{D}_i \in \mathbb{D}$}
    \State $\mathcal{W}_i \leftarrow \mathrm{Segment}\big(\mathcal{D}_i;\, L=f_{s,i},\, \delta=f_{s,i}/2\big),\quad
            \mathbf{X}\in\mathbb{R}^{f_{s,i}\times 6}$
\EndFor
\Statex
\Statex \Comment{\textit{Validation/Test Partitioning} — Sec.~\ref{sec:partitioning}}
\State $\mathcal{W}_i=\mathcal{W}_i^{\mathrm{tr}}\cup\mathcal{W}_i^{\mathrm{val}}\cup\mathcal{W}_i^{\mathrm{te}}$ (disjoint);\quad
       $\mathcal{W}^{\mathrm{val}},\mathcal{W}^{\mathrm{te}}$ fixed
\Statex
\Statex \Comment{\textit{Data Selection} — Sec.~\ref{sec:grid}}
\ForAll{$s \in \mathcal{S}$}
    \For{$n_i \in \mathcal{N}^{s},\; i=1,\dots,10$}
        \For{$r=1,\dots,R$}
            \State $\mathcal{B}\sim\mathcal{W}^{\mathrm{tr}},\;\; |\mathcal{B}|=n_i$
            \State $\theta \leftarrow \theta_0$ \Comment{re-init from scratch}
            \State $\theta^{\ast}\leftarrow \arg\min_{\theta}\,
                   \mathcal{L}_{\mathrm{CE}}(\mathcal{M}_\theta;\mathcal{B})$
                   \Comment{Adam, $\eta=10^{-3}$, $B=64$; early stop on $\mathcal{W}^{\mathrm{val}}$}
            \State $\mathrm{Accuracy}_r(n_i)\leftarrow
                   \mathrm{Accuracy}\big(\mathcal{M}_{\theta^{\ast}};\mathcal{W}^{\mathrm{te}}\big)$
                   \Comment{Eq.~\eqref{eq:softmax_output}}
        \EndFor
        \State $\overline{\mathrm{Accuracy}}(n_i)\leftarrow
               \tfrac{1}{R}\sum_{r=1}^{R}\mathrm{Accuracy}_r(n_i)$
    \EndFor
\EndFor
\State \Return $\big\{(n_i,\,\overline{\mathrm{Accuracy}}(n_i))\big\}_{i=1}^{10}$
\end{algorithmic}
\end{algorithm}

\subsubsection{Data perpetration}\label{sec:windowconstruction}

For each dataset, raw inertial recordings were segmented into overlapping fixed length windows. Given a sampling frequency $f_s$ [Hz], we used a window duration of 1\,second, resulting in $f_s$ samples per window. For example, for PAMAP2, which is sampled at 100\,Hz, each window contains 100 consecutive time steps. To increase the effective number of training examples while preserving strong temporal correlation within windows, we applied a 50\% overlap between consecutive windows. Concretely, with a stride of $f_s/2$ samples.

This procedure yielded a large pool of labeled windows for each dataset and task (binary and multi class), from which we subsequently drew training subsets of controlled sizes.

\subsubsection{Data selection} \label{sec:grid}
During the data selection phase, we captured varying levels of classification difficulty by instantiating each dataset into two distinct scenarios:
\begin{itemize}
     \item \textbf{Binary classification} provide relatively low entropy decision problems (e.g., dynamic vs. static activities), where class boundaries are typically clearer and performance saturates with fewer samples.
    
     \item \textbf{Multi-class classification} impose higher label complexity (e.g., 4–33 activities), where classes may overlap in feature space and rare classes appear, leading to slower and potentially noisier convergence behavior.
\end{itemize}

To study convergence as a function of training set size, we defined a discrete grid of ten measurement points for each scenario. For the binary tasks, we first identified, across all datasets, the smallest available binary window set (i.e., the dataset with the fewest labeled windows after segmentation). Let $N_{\text{bin,min}}$ denote the total number of binary windows in that limiting dataset. We then constructed ten training set sizes as equally spaced fractions of this smallest dataset, e.g.,
\[
10\%,\, \dots,\, 100\% \quad \Rightarrow \quad 10{,}000,\; \dots,\; 100{,}000 \text{ samples}
\]

For the multi class tasks, we applied an analogous procedure with a larger absolute scale, reflecting the greater number of effective classes and the increased sample requirements. Specifically, we defined ten training set sizes such as $35$k, $70$k, \dots, $350$k samples, again anchored to the smallest available multi class dataset. This design ensures that all datasets can be evaluated on the same set of sample size points without extrapolation beyond their available data.

\subsubsection{Validation/Test Partitioning} \label{sec:partitioning}

For each dataset and scenario (binary and multi class), we first created a fixed disjoint partition of the segmented windows into training, validation, and test sets. The validation and test sets were held constant across all sample size conditions. Only the size of the training set was varied along the ten predefined points (e.g., 10k–100k windows for the binary tasks and 35k–350k windows for the multi class tasks), obtained by subsampling from the full training pool.

We trained the same baseline network with identical optimization settings, in order to isolate the effect of training data size from architectural or hyperparameter changes \cite{rajput2023evaluation,silvey2024sample}.

We used the Adam optimizer \cite{kingma2014adam} with an initial learning rate of 0.001 and a batch size of 64 windows. Model optimization was guided by a cross-entropy loss function calculated over the task-specific label set.

The train/validation/test split for each dataset was fixed (Section~\ref{sec:design}), and only the training portion was subsampled to construct the ten sample size points per scenario. At each sample size level, we trained the model from scratch, monitoring validation accuracy for early stopping to avoid overfitting on very small training sets. Test accuracy was evaluated only once per trained model, and all reported results correspond to the mean over multiple random subsampling runs.

By maintaining a constant architecture and training protocol across all experimental configurations, this design ensures that systematic variations in the fitted logarithmic learning curves and classification accuracy are attributable solely to the interplay between training sample size, task granularity, and dataset characteristics, rather than changes in model capacity or optimization details. Since the evaluation distribution remains fixed, this methodology isolates the impact of training data volume, enabling a consistent estimate of the asymptotic accuracy, derived from the complete training set, against which the logarithmic convergence model and the stability thresholds are formally defined.

\subsection{Convergence Analysis} 
Following the model training, the convergence analysis is carried out through a two-phase procedure that transforms the raw training evaluations into a stable empirical learning curve and a quantitative stability estimate.

1. \textbf{Accuracy: }
Model performance was evaluated using classification accuracy on the fixed test set defined in Section~\ref{sec:design}. For each trained model, accuracy was computed as the percentage of test windows whose predicted class matched the corresponding ground truth label for the task specific label set. 
To obtain stable estimates at each sample size condition, every configuration (dataset, task scenario, and training set size) was repeated over multiple random subsampling runs of the training portion. For each run, a separate model was trained and evaluated once on the fixed test set, and the resulting test accuracies were aggregated offline to compute the mean accuracy per configuration, which serves as the basis for the subsequent learning curve fitting.
 
2. \textbf{Curve Fitting and Stability Analysis: } \label{sec:fitting} 
The ten empirical points, evaluated over the sample-size grid defined in Section~\ref{sec:grid}, are used to fit and evaluate logarithmic curves of the form
\begin{equation}
\label{eq:accuracy_model}
\mathrm{Accuracy}(n) = a \log(n) + b
\end{equation}
where $n$ is the number of training samples and $a, b \in \mathbb{R}$ are scenario-specific constants estimated by nonlinear least squares. Because the tenth point corresponds to the full training pool, it does not constitute an independent held-out check and cannot serve as an unbiased reference. The \emph{reference curve} $y^*(n)$ is therefore fitted on the first nine points $\{(n_i, \mathrm{Accuracy}(n_i))\}_{i=1}^{10}$, which represent the widest range of sample sizes available for an unbiased fit.
 
To assess how early a reliable curve can be obtained from a small pilot study, a systematic adequacy analysis is then performed. For each prefix length $N \in \{2, \dots, 10\}$, a \emph{candidate curve} $\hat{y}_N(n)$ is fitted using only the first $N$ points. Both $\hat{y}_N$ and $y^*$ are then evaluated at all ten sample sizes $\{n_1, \dots, n_{10}\}$, and their agreement is quantified by the mean absolute percentage deviation (MAPD):
\begin{equation}
\label{eq:mapd_metric}
\mathrm{MAPD}(\hat{y}_N,\, y^*) = \frac{100}{10} \sum_{i=1}^{10}
\left| \frac{\hat{y}_N(n_i) - y^*(n_i)}{y^*(n_i)} \right|
\end{equation}
The stability point $N^*$ is the smallest $N$ for which the MAPD falls within a predefined tolerance. Because binary tasks yield inherently cleaner decision boundaries and faster convergence, stricter thresholds ($\tau_{\text{lenient}}=2\%$, $\tau_{\text{strict}}=1\%$) are applied to them, while the more complex multi-class tasks use relaxed thresholds ($\tau_{\text{lenient}}=5\%$, $\tau_{\text{strict}}=2\%$). These tolerance levels were fixed prior to data collection based on the expected difficulty of each scenario type. If no candidate curve meets either threshold, the curve with the lowest MAPD is selected as the best available estimate.
 
\section{Analysis and Results}
\label{sec:results}

To produce the results, each dataset is processed through the same standardized pipeline: sliding-window segmentation, training over the incrementally scaled sample-size grid (Section~\ref{sec:grid}), and logarithmic curve fitting with stability analysis based on the MAPD of the candidate curve relative to the reference curve (Section~\ref{sec:fitting}). Then, for each dataset we present its fitted equations, and stability behavior, with full procedural details available in the referenced sections.

Results are reported separately for the binary and multi-class settings of each dataset, focusing on how test accuracy scales with the number of training windows and how quickly the logarithmic model stabilizes as additional sample size points are included in the fitted curves. To maintain an architecture-agnostic perspective, all curves are interpreted primarily as functions of training sample size rather than as evaluations of a particular network design.

For each dataset, the results are presented in two stacked panels. First, a figure plots the observed test accuracy at each sample size together with the logarithmic fits obtained from the first $N \in \{2,\dots,10\}$ sample size points. Here the observed accuracy is the test-set classification accuracy, computed as the percentage of test windows whose predicted label matches the ground truth (GT), i.e., $100 \times (\text{correct predictions} / \text{total predictions})$, averaged over the random subsampling runs, is addressed as baseline. Additionally, for each dataset a table lists the corresponding fitted equations and their MAPD with respect to the reference curve, as defined in Section~\ref{sec:fitting}. Because the performance scale and practical notion of an acceptable deviation differ between binary and multi-class recognition, we use MAPD thresholds of $5\%$ and $2\%$ for the multi-class tasks, and stricter thresholds of $2\%$ and $1\%$ for the binary tasks.

This layout makes it possible to inspect both the qualitative behavior of the curves and the quantitative deviation from the reference as $N$ increases. It also facilitates the identification of the stability point: the smallest $N$ for which the fitted curve achieves a MAPD within the specified tolerance. Identifying this point is critical for optimizing the trade-off between model reliability and the temporal and financial costs of data acquisition, as it indicates the minimum sufficient data required for the model to reach practical convergence.

\subsection{Multi-Class Classification}

For every dataset, the model was trained on the incrementally scaled sample-size grid of Section~\ref{sec:grid} and evaluated on the held-out test set. The observed accuracies were then fitted with the logarithmic model of ~\eqref{eq:accuracy_model_final}, and the MAPD of each fit relative to the reference curve was computed as described in Section~\ref{sec:fitting}. Each dataset is presented as a figure of the learning curves and logarithmic fits, followed by a table of the fitted equations and their MAPD values, with the per-dataset stability behavior discussed below.

\textbf{UCI HAR}: As shown in Figure \ref{fig:uci_har_multi_logfits}, the baseline test accuracy begins near 80\% and scales toward 90\% as training samples increase to 350,000. Table \ref{tab:uci_har_multi_mapd} confirms early stabilization as with only $N=3$ training points (equals to 105,000 samples), the model achieves an MAPD of 0.943\%, satisfying the stricter stability threshold of $<2\%$.

\begin{center}
  \centering
  \includegraphics[width=\linewidth]{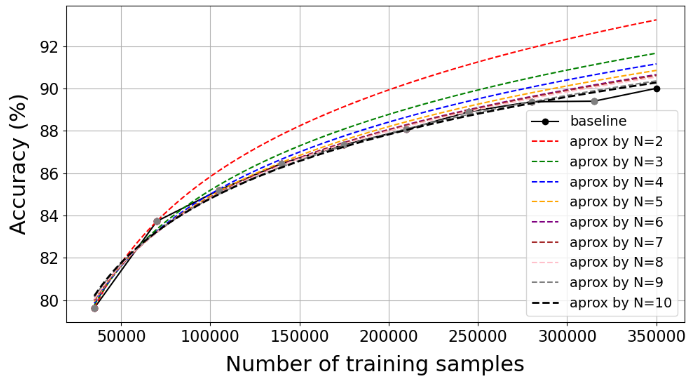}
    \captionof{figure}{UCI HAR (multi class) learning curves with logarithmic fits using the first $N$ sample size points.}
    \label{fig:uci_har_multi_logfits}

  \vspace{0.8em}
  \small
  \begin{tabular}{c l r}
      \toprule
      $N$ & Equation & MAPD (\%) \\
      \midrule
      \rowcolor{gray!30} 
      2  & $5.9 \log(n) + 17.8$      & 2.095 \\
      \rowcolor{gray!10} 
      3  & $5.2 \log(n) + 25.7$      & 0.943 \\
      4  & $4.9 \log(n) + 28.5$      & 0.586 \\
      5  & $4.7 \log(n) + 30.3$      & 0.376 \\
      6  & $4.6 \log(n) + 31.6$      & 0.241 \\
      7  & $4.6 \log(n) + 31.8$      & 0.214 \\
      8  & $4.5 \log(n) + 32.3$      & 0.168 \\
      9  & $4.4 \log(n) + 33.8$      & 0.051 \\
      10 & $4.4\log(n)+34.4$   & 0 \\
      \bottomrule
    \end{tabular}
    \captionof{table}{UCI HAR (multi class) logarithmic fits and MAPD relative to the reference curve. The dark highlight indicates an MAPD $< 5\%$, and the light highlight indicates an MAPD $< 2\%$.}
    \label{tab:uci_har_multi_mapd}
\end{center}

\textbf{Mobile Pos}: This dataset reveals higher initial instability, with Figure \ref{fig:mobile pos_multi_logfits} displaying wide deviations in early logarithmic fits ($N=2$ through $N=5$) that tend to overestimate asymptotic accuracy. Table \ref{tab:Mobile Pos_multi_mapd} highlights this difficulty, as an MAPD below the 5\% threshold is only achieved at $N=7$ (4.091\%), and the stricter 2\% threshold requires $N=9$.

\begin{center}
  \centering
  \includegraphics[width=\linewidth]{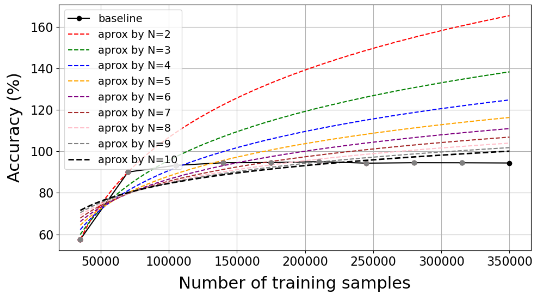}
    \captionof{figure}{Mobile Pos (multi class) learning curves with logarithmic fits using the first $N$ sample size points.}
    \label{fig:mobile pos_multi_logfits}

  \vspace{0.8em}
  \small
  \begin{tabular}{c l r}
      \toprule
      $N$ & Equation & MAPD (\%) \\
      \midrule
      2  & $43.1 \log(n) - 432.8$      & 43.06 \\
      3  & $34.1 \log(n) - 296.4$      & 24.441 \\
      4  & $27.2 \log(n) - 222$      & 15.466 \\
      5  & $22.6 \log(n) - 171.8$      & 10.013 \\
      6  & $19.5 \log(n) - 137.7$      & 6.64 \\
      \rowcolor{gray!30} 
      7  & $16.9 \log(n) - 109.7$      & 4.091 \\
      8  & $15.1 \log(n) - 88.8$      & 2.329 \\
      \rowcolor{gray!10} 
      9  & $13.6 \log(n) - 72.1$      & 1.017 \\
      10 & $12.4 \log(n) - 58.3$   & 0 \\
      \bottomrule
    \end{tabular}
    \captionof{table}{Mobile Pos (multi class) logarithmic fits and MAPD relative to the reference curve. The dark highlight indicates an MAPD $< 5\%$, and the light highlight indicates an MAPD $< 2\%$.}
    \label{tab:Mobile Pos_multi_mapd}
\end{center}

\textbf{MotionSense}: This dataset exhibits high predictability, with Figure \ref{fig:MotionSense_multi_logfits} showing tight clustering of all fits around observed data. Table \ref{tab:MotionSense_multi_mapd} shows that even at $N=2$, the MAPD is 0.471\%, indicating the model reaches the stricter stability threshold from the very first increments.

\begin{center}
  \centering
  \includegraphics[width=\linewidth]{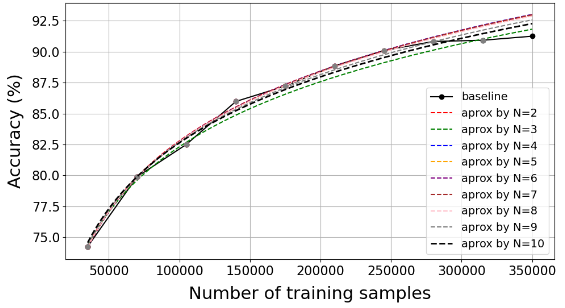}
    \captionof{figure}{MotionSense (multi class) learning curves with logarithmic fits using the first $N$ sample size points.}
    \label{fig:MotionSense_multi_logfits}

  \vspace{0.8em}
  \small
  \begin{tabular}{c l r}
      \toprule
      $N$ & Equation & MAPD (\%) \\
      \midrule
      \rowcolor{gray!10} 
      2  & $8.14 \log(n) - 10.9$      & 0.471 \\
      3  & $7.6 \log(n) - 5$      & 0.43 \\
      4  & $8.2 \log(n) - 11.1$      & 0.472 \\
      5  & $8.1 \log(n) - 11.1$      & 0.417 \\
      6  & $8.2 \log(n) - 11.4$      & 0.45 \\
      7  & $8.2 \log(n) - 10.7$      & 0.459 \\
      8  & $8.1 \log(n) - 10.7$      & 0.386 \\
      9  & $7.9 \log(n) - 8.3$      & 0.192 \\
      10 & $7.7 \log(n) - 5.9$   & 0 \\
      \bottomrule
    \end{tabular}
    \captionof{table}{MotionSense (multi class) logarithmic fits and MAPD relative to the reference curve. The dark highlight indicates an MAPD $< 5\%$, and the light highlight indicates an MAPD $< 2\%$.}
    \label{tab:MotionSense_multi_mapd}
\end{center}

\textbf{WISDM}: The results show a steady incline in accuracy from approximately 57\% to 67\% in Figure \ref{fig:WISDM_multi_logfits}. While early points show low deviation, Table \ref{tab:WISDM_multi_mapd} indicates that consistent stability within the 5\% threshold is met at $N=4$ (2.147\%), and the stricter 2\% threshold is reached and maintained starting at $N=6$ (1.314\%).

\begin{center}
  \centering
  \includegraphics[width=\linewidth]{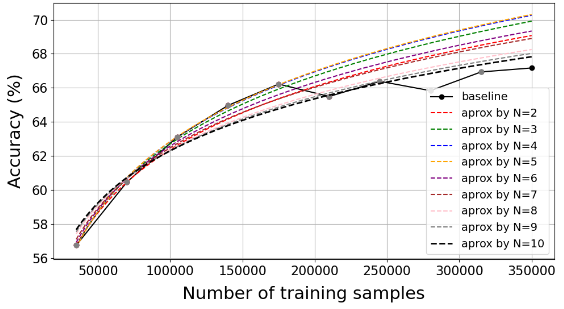}
    \captionof{figure}{WISDM (multi class) learning curves with logarithmic fits using the first $N$ sample size points.}
    \label{fig:WISDM_multi_logfits}

  \vspace{0.8em}
  \small
  \begin{tabular}{c l r}
      \toprule
      $N$ & Equation & MAPD (\%) \\
      \midrule
      2  & $5.3 \log(n) + 0.8$      & 1.05 \\
      3  & $5.7 \log(n) - 3.5$      & 1.825 \\
      \rowcolor{gray!30} 
      4  & $5.9 \log(n) - 5.4$      & 2.147 \\
      5  & $6 \log(n) - 5.7$      & 2.194 \\
      \rowcolor{gray!10} 
      6  & $5.4 \log(n) + 3.5$      & 1.314 \\
      7  & $5.2 \log(n) + 8.2$      & 0.928 \\
      8  & $4.7 \log(n) + 8.2$      & 0.363 \\
      9  & $4.6 \log(n) + 9.9$      & 0.168 \\
      10 & $4.4 \log(n)+ 11.5$   & 0 \\
      \bottomrule
    \end{tabular}
    \captionof{table}{WISDM (multi class) logarithmic fits and MAPD relative to the reference curve. The dark highlight indicates an MAPD $< 5\%$, and the light highlight indicates an MAPD $< 2\%$.}
    \label{tab:WISDM_multi_mapd}
\end{center}

\textbf{PAMAP2}: This dataset presents a more challenging scenario due to the high number of class. In Figure \ref{fig:PAMAP2_multi_logfits}, initial fits deviate significantly from the observed mean. Table \ref{tab:PAMAP2_multi_mapd} shows the 5\% MAPD threshold is satisfied at $N=7$ (3.804\%), while universal stability at the stricter 2\% level is reached at $N=9$.

\begin{center}
  \centering
  \includegraphics[width=\linewidth]{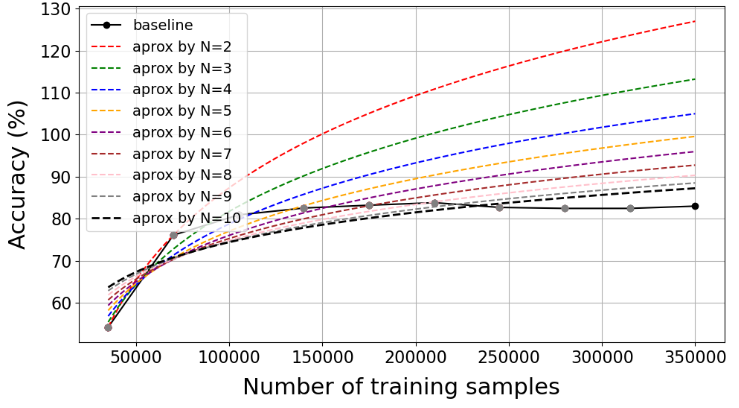}
    \captionof{figure}{PAMAP2 (multi class) learning curves with logarithmic fits using the first $N$ sample size points.}
    \label{fig:PAMAP2_multi_logfits}

  \vspace{0.8em}
  \small
  \begin{tabular}{c l r}
      \toprule
      $N$ & Equation & MAPD (\%) \\
      \midrule
      
      2  & $31.6 \log(n) - 276.5$      & 29.655 \\
      3  & $25.1 \log(n) - 207.2$      & 18.879 \\
      4  & $20.9 \log(n) - 162.1$      & 12.659 \\
      5  & $18 \log(n) - 130$      & 8.683 \\
      6  & $15.9 \log(n) - 107$      & 6.083 \\
      \rowcolor{gray!30} 
      7  & $13.9 \log(n) - 85.1$      & 3.804 \\
      8  & $12.4 \log(n) - 67.6$      & 2.113 \\
      \rowcolor{gray!10} 
      9  & $11.1 \log(n) - 53.7$      & 0.866 \\
      10 & $10.2 \log(n) - 43.4$   & 0 \\
      \bottomrule
    \end{tabular}
    \captionof{table}{PAMAP2 (multi class) logarithmic fits and MAPD relative to the reference curve. The dark highlight indicates an MAPD $< 5\%$, and the light highlight indicates an MAPD $< 2\%$.}
    \label{tab:PAMAP2_multi_mapd}
\end{center}

\textbf{REALDISP}: This dataset, containing the highest number of classes (33 activities), displays a well-behaved learning curve in Figure \ref{fig:REALDISP multi_logfits}. According to Table \ref{tab:REALDISP_multi_mapd}, the model reaches consistent 5\% tolerance starting at $N=3$ (4.579\%) and the stricter 2\% tolerance by $N=5$ (1.727\%).  

\begin{center}
  \centering
  \includegraphics[width=\linewidth]{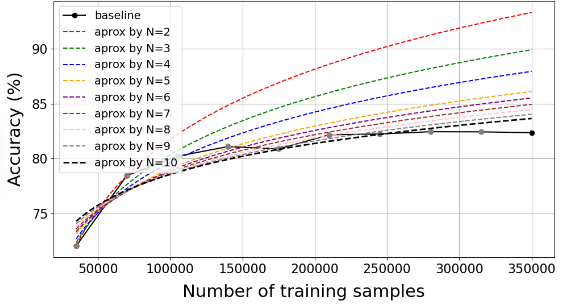}
    \captionof{figure}{REALDISP (multi class) learning curves with logarithmic fits using the first $N$ sample size points.}
    \label{fig:REALDISP multi_logfits}

  \vspace{0.8em}
  \small
  \begin{tabular}{c l r}
      \toprule
      $N$ & Equation & MAPD (\%) \\
      \midrule
      
      2  & $9.21 \log(n) - 24.4$      & 7.254 \\
      \rowcolor{gray!30} 
      3  & $7.6 \log(n) - 7.3$      & 4.579 \\
      4  & $6.6 \log(n) + 3.5$      & 3.086 \\
      \rowcolor{gray!10} 
      5  & $5.6 \log(n) + 14.4$      & 1.727 \\
      6  & $5.3 \log(n) + 18.1$      & 1.304 \\
      7  & $4.9 \log(n) + 22.1$      & 0.884 \\
      8  & $4.6 \log(n) + 25.5$      & 0.557 \\
      9  & $4.3 \log(n) + 28.8$      & 0.263 \\
      10 & $4.1 \log(n) + 31.9$   & 0 \\
      \bottomrule
    \end{tabular}
    \captionof{table}{REALDISP (multi class) logarithmic fits and MAPD relative to the reference curve. The dark highlight indicates an MAPD $< 5\%$, and the light highlight indicates an MAPD $< 2\%$.}
    \label{tab:REALDISP_multi_mapd}
\end{center}

Across all six datasets, the multi-class results exhibit the same qualitative behavior: test accuracy rises rapidly for small training sets and then saturates, and in every case this growth is well captured by the logarithmic model of Eq.~\eqref{eq:accuracy_model_final}, with the MAPD relative to the reference curve decreasing monotonically as additional sample-size points are included. Datasets with fewer, well-separated classes (such as MotionSense and UCI HAR) stabilize almost immediately, whereas the higher-granularity datasets (PAMAP2 and Mobile Pos) require more points before reaching the same tolerance.
\subsection{Binary Classification}

We now turn to the binary setting, in which the activity labels of each dataset are collapsed into two classes. The same procedure is applied as in the multi-class case: the network is trained over the sample-size grid of Section~\ref{sec:grid}, the resulting test accuracies are fitted with the logarithmic model of Eq.~\eqref{eq:accuracy_model_final}, and the MAPD of each fit relative to the reference curve is computed as defined in Section~\ref{sec:fitting}. As before, every dataset is shown as a learning-curve figure followed by a table of fitted equations and MAPD values. Because the binary task operates at a higher and more compressed accuracy range, stricter MAPD thresholds of $2\%$ and $1\%$ are used to assess stability.

\textbf{UCI HAR}: Figure \ref{fig:uci_har_binary_logfits} illustrates the trajectory of accuracy growth for the binary task. According to Table \ref{tab:uci_har_binary_mapd}, the model meets the 2\% stability requirement by $N=4$ (1.884\%) and reaches the stricter 1\% threshold by $N=7$.

\begin{center}
  \centering
  \includegraphics[width=\linewidth]{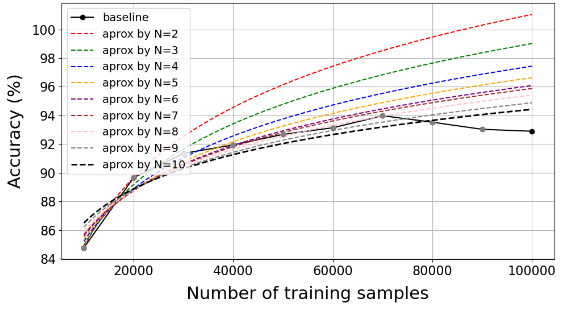}
    \captionof{figure}{UCI HAR (binary) learning curves with logarithmic fits using the first $N$ sample size points.}
    \label{fig:uci_har_binary_logfits}

  \vspace{0.8em}
  \small
  \begin{tabular}{c l r}
      \toprule
      $N$ & Equation & MAPD (\%) \\
      \midrule
      2  & $7.1 \log(n) + 19.7$      & 4.339 \\
      3  & $6.1 \log(n) + 28.7$      & 2.939 \\
      \rowcolor{gray!30} 
      4  & $5.3 \log(n) + 36.4$      & 1.884 \\
      5  & $4.9 \log(n) + 40.6$      & 1.363 \\
      6  & $4.5 \log(n) + 43.7$      & 1.013 \\
      \rowcolor{gray!10} 
      7  & $4.4 \log(n) + 44.9$      & 0.886 \\
      8  & $4.1 \log(n) + 47.9$      & 0.602 \\
      9  & $3.8 \log(n) + 51.6$      & 0.275 \\
      10 & $3.4\log(n)+54.8$   & 0 \\
      \bottomrule
    \end{tabular}
    \captionof{table}{UCI HAR (binary) logarithmic fits and MAPD relative to the reference curve. The dark highlight indicates an MAPD $< 2\%$, and the light highlight indicates an MAPD $< 1\%$.}
    \label{tab:uci_har_binary_mapd}
\end{center}

\textbf{Mobile Pos}: The binary task stabilizes almost immediately, as seen in Figure \ref{fig:mobile pos_binary_logfits}. Table \ref{tab:Mobile Pos_binary_mapd} shows that by $N=3$ (equals to 30,000 samples), the MAPD is 0.992\%, satisfying the stricter $<1$\% tolerance level.

\begin{center}
  \centering
  \includegraphics[width=\linewidth]{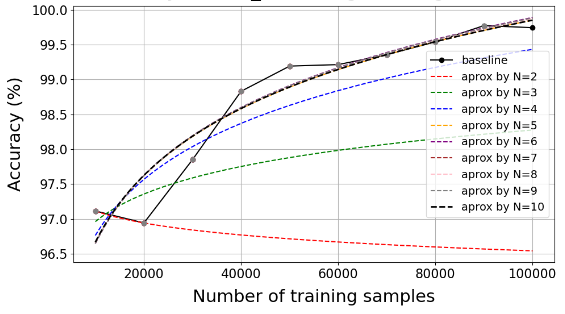}
    \captionof{figure}{Mobile Pos (binary) learning curves with logarithmic fits using the first $N$ sample size points.}
    \label{fig:mobile pos_binary_logfits}

  \vspace{0.8em}
  \small
  \begin{tabular}{c l r}
      \toprule
      $N$ & Equation & MAPD (\%) \\
      \midrule
      2  & $ 0.2 \log(n) + 99.4$      & 2.126 \\
      \rowcolor{gray!30} 
      3  & $0.57 \log(n) + 91.7$      & 0.992 \\
      \rowcolor{gray!10} 
      4  & $1.15 \log(n) + 86.1$      & 0.259 \\
      5  & $1.4 \log(n) + 83.9$      & 0.003 \\
      6  & $1.41 \log(n) + 83.7$      & 0.002 \\
      7  & $1.4 \log(n) + 83.75$      & 0.015 \\
      8  & $1.4 \log(n) + 83.8$      & 0.01 \\
      9  & $1.4 \log(n) + 83.7$      & 0.018 \\
      10 & $1.4 \log(n)+ 83.9$   & 0 \\
      \bottomrule
    \end{tabular}
    \captionof{table}{Mobile Pos (binary) logarithmic fits and MAPD relative to the reference curve. The dark highlight indicates an MAPD $< 2\%$, and the light highlight indicates an MAPD $< 1\%$.}
    \label{tab:Mobile Pos_binary_mapd}
\end{center}

\textbf{MotionSense}: This dataset reflects a smooth growth pattern in Figure \ref{fig:MotionSense_binary_logfits}. As detailed in Table \ref{tab:MotionSense_binary_mapd}, the scenario meets the 2\% threshold at $N=4$ (1.742\%) and the stricter 1\% threshold by $N=5$ (0.961\%).

\begin{center}
  \centering
  \includegraphics[width=\linewidth]{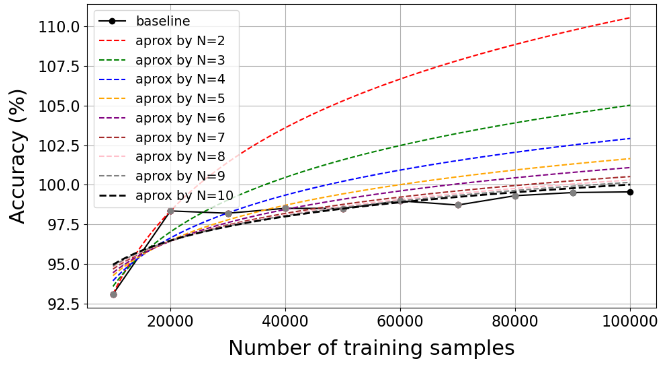}
    \captionof{figure}{MotionSense (binary) learning curves with logarithmic fits using the first $N$ sample size points.}
    \label{fig:MotionSense_binary_logfits}

  \vspace{0.8em}
  \small
  \begin{tabular}{c l r}
      \toprule
      $N$ & Equation & MAPD (\%) \\
      \midrule
      2  & $7.6 \log(n) + 23.2$      & 6.686 \\
      3  & $4.9 \log(n) + 47.8$      & 3.068 \\
      \rowcolor{gray!30} 
      4  & $3.9 \log(n) + 58$      & 1.742 \\
      \rowcolor{gray!10} 
      5  & $3.2 \log(n) +64.7$      & 0.961 \\
      6  & $2.9 \log(n) + 67.8$      & 0.622 \\
      7  & $2.5 \log(n) + 71.3$      & 0.289 \\
      8  & $2.4 \log(n) + 72.7$      & 0.163 \\
      9  & $2.3 \log(n) + 73.7$      & 0.078 \\
      10 & $2.2 \log(n)+ 74.7$   & 0 \\
      \bottomrule
    \end{tabular}
    \captionof{table}{MotionSense (binary) logarithmic fits and MAPD relative to the reference curve. The dark highlight indicates an MAPD $< 2\%$, and the light highlight indicates an MAPD $< 1\%$.}
    \label{tab:MotionSense_binary_mapd}
\end{center}

\textbf{WISDM}: The learning trajectory is shown in Figure \ref{fig:WISDM_binary_logfits}. Table \ref{tab:WISDM_binary_mapd} shows the model reaches 2\% stability at $N=6$ (1.997\%) but requires $N=8$ points to fall below the stricter 1\% tolerance mark (0.627\%).

\begin{center}
  \centering
  \includegraphics[width=\linewidth]{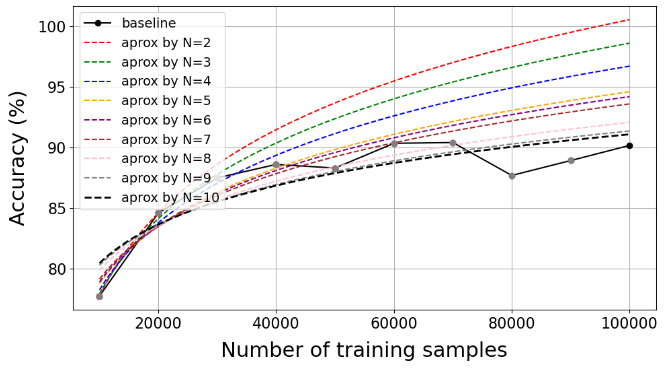}
    \captionof{figure}{WISDM (binary) learning curves with logarithmic fits using the first $N$ sample size points.}
    \label{fig:WISDM_binary_logfits}

  \vspace{0.8em}
  \small
  \begin{tabular}{c l r}
      \toprule
      $N$ & Equation & MAPD (\%) \\
      \midrule
      2  & $9.9 \log(n) - 13.4$      & 6.42 \\
      3  & $9 \log(n) - 4.8$      & 5.028 \\
      4  & $8 \log(n) + 4.3$      & 3.702 \\
      5  & $6.9 \log(n) + 15.4$      & 2.67 \\
      \rowcolor{gray!30} 
      6  & $6.6 \log(n) + 17.7$      & 1.997 \\
      7  & $6.3 \log(n) + 21.3$      & 1.605 \\
      \rowcolor{gray!10} 
      8  & $5.3 \log(n) + 31.1$      & 0.627 \\
      9  & $4.8 \log(n) + 36$      & 0.172 \\
      10 & $4.6 \log(n)+ 37.9$   & 0 \\
      \bottomrule
    \end{tabular}
    \captionof{table}{WISDM (binary) logarithmic fits and MAPD relative to the reference curve. The dark highlight indicates an MAPD $< 2\%$, and the light highlight indicates an MAPD $< 1\%$.}
    \label{tab:WISDM_binary_mapd}
\end{center}

\textbf{PAMAP2}: The learning curve for this binary task is displayed in Figure \ref{fig:PAMAP2_binary_logfits}. Table \ref{tab:PAMAP2_binary_mapd} indicates that it requires $N=7$ points to reach the 2\% threshold (1.97\%) and $N=9$ to reach the stricter 1\% threshold (0.479\%).

\begin{center}
  \centering
  \includegraphics[width=\linewidth]{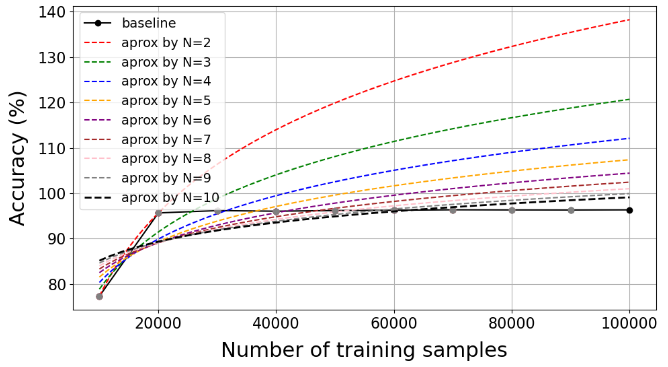}
    \captionof{figure}{PAMAP2 (binary) learning curves with logarithmic fits using the first $N$ sample size points.}
    \label{fig:PAMAP2_binary_logfits}

  \vspace{0.8em}
  \small
  \begin{tabular}{c l r}
      \toprule
      $N$ & Equation & MAPD (\%) \\
      \midrule
      
      2  & $26.4 \log(n) - 166.3$      & 25.316 \\
      3  & $18.2 \log(n) - 88.38$      & 13.536 \\
      4  & $13.8 \log(n) - 46.7$      & 7.962 \\
      5  & $11.2 \log(n) - 21.9$      & 4.988 \\
      6  & $9.5 \log(n) - 5.17$      & 3.168 \\
      \rowcolor{gray!30} 
      7  & $8.3 \log(n) + 6.78$      & 1.97 \\
      8  & $7.4 \log(n) + 16$      & 1.113 \\
      \rowcolor{gray!10} 
      9  & $6.6 \log(n) + 23.4$      & 0.479 \\
      10 & $6 \log(n) + 29.3$   & 0 \\
      \bottomrule
    \end{tabular}
    \captionof{table}{PAMAP2 (binary) logarithmic fits and MAPD relative to the reference curve. The dark highlight indicates an MAPD $< 2\%$, and the light highlight indicates an MAPD $< 1\%$.}
    \label{tab:PAMAP2_binary_mapd}
\end{center}

\textbf{REALDISP}: Figure \ref{fig:REALDISP binary_logfits} exhibits a very flat logarithmic curve, indicating near-maximum accuracy is reached with minimal samples. Table \ref{tab:REALDISP_binary_mapd} confirms this, with an MAPD of 0.159\% at $N=2$, satisfying both the 2\% and the stricter 1\% thresholds immediately.

\begin{center}
  \centering
  \includegraphics[width=\linewidth]{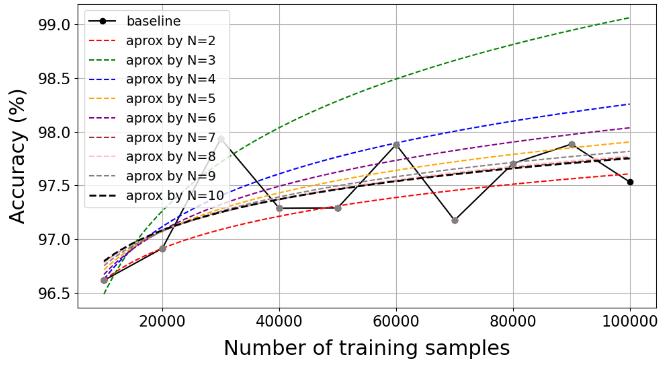}
    \captionof{figure}{REALDISP (binary) learning curves with logarithmic fits using the first $N$ sample size points.}
    \label{fig:REALDISP binary_logfits}

  \vspace{0.8em}
  \small
  \begin{tabular}{c l r}
      \toprule
      $N$ & Equation & MAPD (\%) \\
      \midrule
      \rowcolor{gray!10} 
      2  & $0.4 \log(n) +92.6$      & 0.159 \\
      3  & $1.1 \log(n) +86.2$      & 0.825 \\
      4  & $0.7 \log(n) + 90.1$      & 0.308 \\
      5  & $0.5 \log(n) + 91.9$      & 0.089 \\
      6  & $0.6 \log(n) + 91.2$      & 0.169 \\
      7  & $0.4 \log(n) + 92.9$      & 0.006 \\
      8  & $0.4 \log(n) + 92.8$      & 0.013 \\
      9  & $0.5 \log(n) + 92.5$      & 0.038 \\
      10 & $0.4 \log(n) + 92$   & 0 \\
      \bottomrule
    \end{tabular}
    \captionof{table}{REALDISP (binary) logarithmic fits and MAPD relative to the reference curve. The dark highlight indicates an MAPD $< 2\%$, and the light highlight indicates an MAPD $< 1\%$.}
    \label{tab:REALDISP_binary_mapd}
\end{center}

The binary results follow the same logarithmic pattern across all six datasets, with convergence that is consistently faster owing to the simpler decision boundary: accuracy saturates after only a few sample-size increments, and the MAPD again decreases monotonically with $N$. The most separable datasets (REALDISP and Mobile Pos) stabilize almost immediately, while the most demanding (PAMAP2 and WISDM) require additional points before reaching the same tolerance.

\subsection{Ablation Study} \label{sec:Sensitivity}
In order to further demonstrate the generalization of our results, we conducted additional experiments utilizing a three-layer CNN architecture, the findings of which are detailed in the following figures and tables.

\textbf{UCI HAR (3-layer CNN)}: The results in Figure~\ref{fig:uci multi claa with CNN log} and Table~\ref{tab:uci multi claa with CNN_mapd} were produced by repeating the procedure of the previous subsections on the UCI HAR multi-class task while replacing the baseline network with the three-layer CNN architecture: the CNN was trained on progressively larger random subsamples at the ten sample-size points of Section~\ref{sec:grid}, and the resulting test accuracies were fitted with the logarithmic model of Eq.~\eqref{eq:accuracy_model_final} and evaluated by MAPD relative to the reference curve.

\begin{center}
  \centering
  \includegraphics[width=\linewidth]{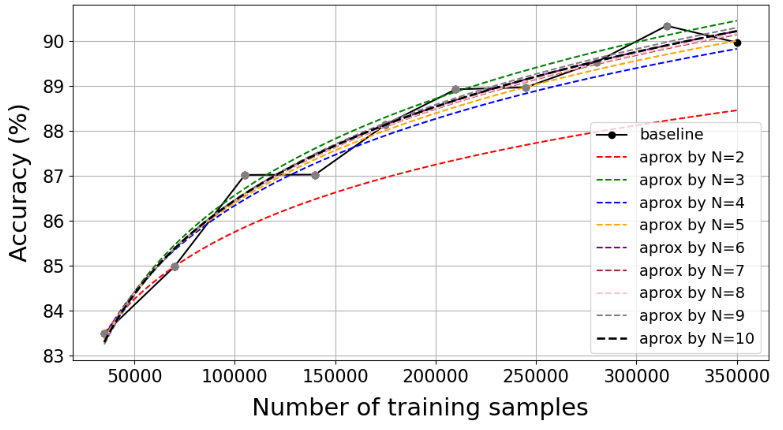}
    \captionof{figure}{UCI-HAR (multi class) learning curves with logarithmic fits using the first $N$ sample size points, for 3 layer CNN.}
    \label{fig:uci multi claa with CNN log}

  \vspace{0.8em}
  \small
  \begin{tabular}{c l r}
      \toprule
      $N$ & Equation & MAPD (\%) \\
      \midrule
      \rowcolor{gray!10} 
      2  & $2.1 \log(n) +60.9$      & 1.279 \\
      3  & $3.1 \log(n) +50.9$      & 0.182 \\
      4  & $2.8 \log(n) + 54.4$      & 0.266 \\
      5  & $2.9 \log(n) + 53.3$      & 0.143 \\
      6  & $3 \log(n) + 52$      & 0.002 \\
      7  & $2.9 \log(n) + 52.5$      & 0.049 \\
      8  & $3 \log(n) + 52.3$      & 0.039 \\
      9  & $3.1 \log(n) + 51.3$      & 0.048 \\
      10 & $3 \log(n) + 52$   & 0 \\
      \bottomrule
    \end{tabular}
    \captionof{table}{UCI-HAR (multi class) logarithmic fits and MAPD relative to the reference curve. The dark highlight indicates an MAPD $< 5\%$, and the light highlight indicates an MAPD $< 2\%$.}
    \label{tab:uci multi claa with CNN_mapd}
\end{center}

As illustrated in \ref{fig:uci multi claa with CNN log}, the accuracy growth for the 3 layer CNN model follows the same consistent logarithmic trajectory as the training set increases toward 350,000 samples. \ref{tab:uci multi claa with CNN_mapd} demonstrates that the model reaches a high degree of stability almost immediately; at $N=2$, the MAPD is already 1.279\%, satisfying the stricter 2\% threshold.

\textbf{PAMAP2 (3-layer CNN)}: The results in Figure~\ref{fig:PAMAP2 multi claa with CNN log} and Table~\ref{tab:PAMAP2 multi class with CNN_mapd} were produced by applying the same three-layer CNN procedure to the PAMAP2 multi-class task: the CNN was trained on progressively larger random subsamples at the ten sample-size points of Section~\ref{sec:grid}, and the resulting test accuracies were fitted with the logarithmic model of Eq.~\eqref{eq:accuracy_model_final} and evaluated by MAPD relative to the reference curve.

\begin{center}
  \centering
  \includegraphics[width=\linewidth]{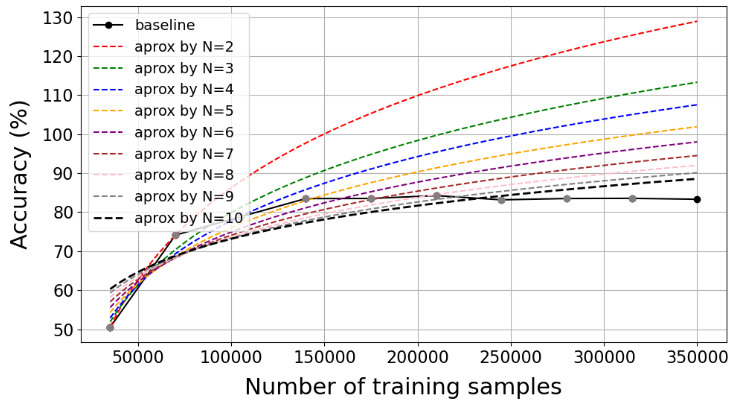}
    \captionof{figure}{PAMAP2 (multi class) learning curves with logarithmic fits using the first $N$ sample size points, for 3 layer CNN.}
    \label{fig:PAMAP2 multi claa with CNN log}

  \vspace{0.8em}
  \small
  \begin{tabular}{c l r}
      \toprule
      $N$ & Equation & MAPD (\%) \\
      \midrule
      
      2  & $34.4 \log(n) -305.7$      & 30.016 \\
      3  & $26.7 \log(n) -277.1$      & 17.843 \\
      4  & $23.7 \log(n) -195.3$      & 13.496 \\
      5  & $20.7 \log(n) -161.99$      & 9.374 \\
      6  & $18.4 \log(n) -137.2$      & 6.578 \\
      \rowcolor{gray!30} 
      7  & $16.3 \log(n) -113.4$      & 4.109 \\
      8  & $14.7 \log(n) -95.4$      & 2.379 \\
      \rowcolor{gray!10} 
      9  & $13.4 \log(n) -80.7$      & 1.065 \\
      10 & $12.8 \log(n) -68.1$   & 0 \\
      \bottomrule
    \end{tabular}
    \captionof{table}{PAMAP2 (multi class) logarithmic fits and MAPD relative to the reference curve. The dark highlight indicates an MAPD $< 5\%$, and the light highlight indicates an MAPD $< 2\%$.}
    \label{tab:PAMAP2 multi class with CNN_mapd}
\end{center}

The learning curve in \ref{fig:PAMAP2 multi claa with CNN log} shows greater initial deviation compared to UCI HAR, which is typical for this high-granularity dataset. However, \ref{tab:PAMAP2 multi class with CNN_mapd} shows that the logarithmic fits converge steadily, satisfying the 5\% MAPD threshold by $N=7$ (4.109\%) and reaching the stricter 2\% stability requirement by $N=9$ (1.065\%).

To assess the generalizability of the proposed framework, an initial extension of this work examined whether the logarithmic learning curve behavior persists when utilizing a different model architecture. We evaluated a three-layer CNN on the UCI HAR and PAMAP2 multi-class tasks, as detailed in \ref{fig:uci multi claa with CNN log} and \ref{fig:PAMAP2 multi claa with CNN log}. The results yielded qualitatively similar patterns to the baseline architecture across all training sample sizes. As shown in \ref{tab:uci multi claa with CNN_mapd} and \ref{tab:PAMAP2 multi class with CNN_mapd}, the CNN models achieved consistent stability thresholds within similar sample size increments as the original BiLSTM-based network. Specifically, UCI HAR reached the 5\% MAPD threshold by $N=2$ (1.279\%), while the more complex PAMAP2 dataset achieved this stability by $N=7$ (4.109\%). This observation suggests that the proposed analysis is robust to changes in model complexity, indicating that the dominant factor in the observed learning trajectories is the underlying structure and volume of the training data rather than the specific choice of neural architecture. Nevertheless, the experiments in this study were intentionally designed to isolate the impact of training set size. The primary contribution remains a data centric framework intended to assist researchers in planning and interpreting sample size requirements, rather than a claim regarding optimal network design for human activity recognition.

\subsection{Summary}
\label{sec:Discussion}
Figures \ref{fig:multi} and \ref{fig:binary} present a meta analysis of the convergence behavior for the multi class and binary scenarios, respectively. These plots quantify the curve fit stability via the stability point $N^*$, the smallest number of training increments ($N$) required for the logarithmic model ($\mathrm{Accuracy} = a \log(n) + b$) to accurately predict the full learning curve within a specified MAPD. Overall, the aggregated results demonstrate that only a modest number of sample size points is required to obtain a stable logarithmic learning curve for most datasets. For multi class tasks, $4/6$ datasets satisfy the $5\%$ MAPD tolerance by $N \approx 4$, with universal stability within a $2\%$ tolerance reached by $N=9$. For binary tasks, convergence is accelerated: $4/6$ datasets meet the $2\%$ MAPD threshold around $N=4$, and all fall within the stringent $1\%$ tolerance by $N=9$. These findings suggest that researchers can reliably extrapolate the asymptotic accuracy of a dataset using our proposed empirical framework after only a few pilot training runs, reducing the burden of massive recording campaigns that are complex, expressive, and time-consuming.
 
\begin{center}
\includegraphics[width=\linewidth]{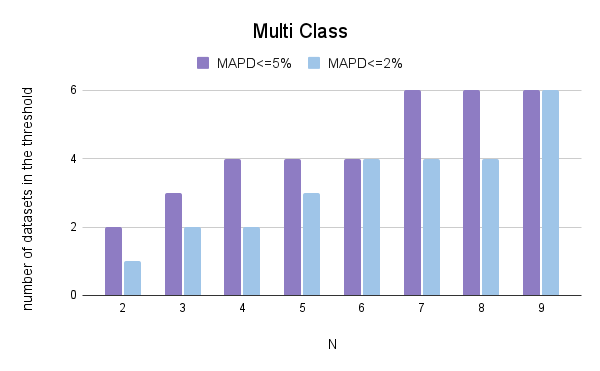}
\captionof{figure}{Cumulative stability thresholds across datasets (multi-class scenario).}
\label{fig:multi}
\end{center}
 
\begin{center}
\includegraphics[width=\linewidth]{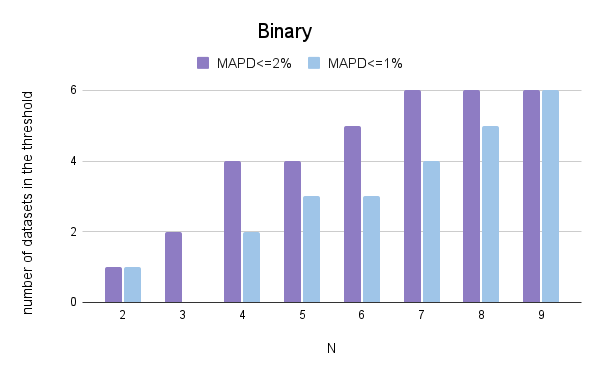}
\captionof{figure}{Cumulative stability thresholds across datasets (binary scenario).}
\label{fig:binary}
\end{center}


\section{Conclusions}
\label{sec:conclusions}
Despite the widespread adoption of deep learning in sensor-based domains, data-driven guidelines for determining the minimum training-set size sufficient to achieve robust classification performance are still unavailable. To address this gap, we proposed a unified empirical framework that characterizes learning-curve convergence as a consistent logarithmic growth in accuracy with respect to training-set size. Its central advantage is that it lets researchers quantitatively define a stability point, the sample size beyond which performance remains within a predefined tolerance of the asymptotic maximum, and thereby extrapolate a task's total data requirements from only a few small-scale pilot runs. Because it relies on a small set of incrementally scaled training increments, the method is inexpensive to apply and readily transferable across datasets and tasks.
 
Applied across six diverse HAR and SLR datasets, the framework revealed that every case follows the proposed logarithmic regime regardless of task complexity, in both the binary and multi-class settings. In the multi-class scenario the majority of datasets reached a five percent stability tolerance after only about four training increments, while in the binary scenario the same proportion satisfied a stricter two percent tolerance at that point; in both cases, universal stability across all datasets was reached within nine increments even under the most stringent tolerances. The same behavior persisted when the baseline recurrent network was replaced with a convolutional architecture, indicating that the observed trajectories are governed primarily by the structure and volume of the training data rather than by the specific model. We conclude that classification performance in bounded sensor based tasks does not follow the unbounded power-law scaling frequently reported in vision and language research, but instead saturates in a predictable logarithmic fashion. This predictability is practically significant, since the asymptotic accuracy of a dataset can be reliably estimated from a handful of incrementally scaled training runs before committing to large-scale data collection. The consistency of the pattern across datasets of differing scale, class granularity, and sensor configuration suggests that it reflects a general property of bounded inertial-sensing classification rather than an artifact of any single dataset, and its robustness to a change of network architecture further indicates that the effect is driven by the data rather than by modeling choices.

Our approach main limitation is that it presumes accuracy grows logarithmically. Although this held consistently across the HAR and SLR tasks we evaluated, it remains an empirical assumption rather than a guarantee for other tasks.
 
While the proposed framework consistently demonstrates logarithmic convergence across diverse datasets, task types, and architectures, several limitations exist: the framework is formulated around saturating accuracy metrics and was validated only on classification tasks, so its extension to regression based models, remains to be established. Even so, the importance of our results lies in their ability to provide concrete, data-backed guidelines for the engineering of ubiquitous computing systems. From a specific perspective, the framework enables HAR and SLR researchers to avoid unnecessary massive recording campaigns that are complex, expressive, and time-consuming, by identifying the point of diminishing returns and directing recording effort only where it still improves performance. From a global perspective, it shifts the prevailing deep-learning objective from maximizing data volume toward optimizing data efficiency. This reframing supports more sustainable and cost-effective development of intelligent sensor based applications, and offers a transferable methodology that other bounded-accuracy domains can adopt to plan their own data collection.

\bibliographystyle{cas-model2-names}
\bibliography{main}

@inproceedings{cortes1993learning,
  author    = {Cortes, Corinna and Jackel, L. D. and Solla, Sara A. and Vapnik, Vladimir and Denker, John S.},
  title     = {Learning curves: Asymptotic values and rate of convergence},
  booktitle = {Advances in Neural Information Processing Systems},
  year      = {1993},
  pages     = {327--334},
  publisher = {Morgan Kaufmann}
}

@article{hutter2021learning,
  author  = {Hutter, Marcus},
  title   = {Learning curve theory},
  journal = {arXiv preprint arXiv:2102.04074},
  year    = {2021}
}

@article{rajput2023evaluation,
  title={Evaluation of a decided sample size in machine learning applications},
  author={Rajput, Daniyal and Wang, Wei-Jen and Chen, Chun-Chuan},
  journal={BMC bioinformatics},
  volume={24},
  number={1},
  pages={48},
  year={2023},
  publisher={Springer}
}

@article{melvin2021sample,
  title={Sample size in machine learning and artificial intelligence},
  author={Melvin, Ryan L},
  journal={Uab. edu. Available online: https://sites. uab. edu/periop-datascience/2021/06/28/sample-size-in-machine-learning-and-artificial-intelligence/(accessed on 30 May 2022)},
  year={2021}
}

@article{rosenfeld2019constructive,
  title={A constructive prediction of the generalization error across scales},
  author={Rosenfeld, Jonathan S and Rosenfeld, Amir and Belinkov, Yonatan and Shavit, Nir},
  journal={arXiv preprint arXiv:1909.12673},
  year={2019}
}

@article{dhiman2023sample,
  title={Sample size requirements are not being considered in studies developing prediction models for binary outcomes: a systematic review},
  author={Dhiman, Paula and Ma, Jie and Qi, Cathy and Bullock, Garrett and Sergeant, Jamie C and Riley, Richard D and Collins, Gary S},
  journal={BMC Medical Research Methodology},
  volume={23},
  number={1},
  pages={188},
  year={2023},
  publisher={Springer}
}

@article{silvey2024sample,
  title={Sample size requirements for popular classification algorithms in tabular clinical data: empirical study},
  author={Silvey, Scott and Liu, Jinze},
  journal={Journal of Medical Internet Research},
  volume={26},
  pages={e60231},
  year={2024},
  publisher={JMIR Publications Toronto, Canada}
}

@article{loog2022survey,
  title={A survey of learning curves with bad behavior: or how more data need not lead to better performance},
  author={Loog, Marco and Viering, Tom},
  journal={arXiv preprint arXiv:2211.14061},
  year={2022}
}

@misc{pamap2_physical_activity_monitoring_231,
  author       = {Reiss, Attila},
  title        = {{PAMAP2 Physical Activity Monitoring}},
  year         = {2012},
  howpublished = {UCI Machine Learning Repository},
  note         = {{DOI}: https://doi.org/10.24432/C5NW2H}
}

@inproceedings{yan2018ridi,
  title={RIDI: Robust IMU double integration},
  author={Yan, Hang and Shan, Qi and Furukawa, Yasutaka},
  booktitle={Proceedings of the European conference on computer vision (ECCV)},
  pages={621--636},
  year={2018}
}

@inproceedings{malekzadeh2019mobile,
  title={Mobile sensor data anonymization},
  author={Malekzadeh, Mohammad and Clegg, Richard G and Cavallaro, Andrea and Haddadi, Hamed},
  booktitle={Proceedings of the international conference on internet of things design and implementation},
  pages={49--58},
  year={2019}
}

@inproceedings{uddin2015human,
  title={Human activity recognition from wearable sensors using extremely randomized trees},
  author={Uddin, Md Taufeeq and Uddiny, Md Azher},
  booktitle={2015 International conference on electrical engineering and information communication technology (ICEEICT)},
  pages={1--6},
  year={2015},
  organization={IEEE}
}

@article{sze2017efficient,
  title={Efficient processing of deep neural networks: A tutorial and survey},
  author={Sze, Vivienne and Chen, Yu-Hsin and Yang, Tien-Ju and Emer, Joel S},
  journal={Proceedings of the IEEE},
  volume={105},
  number={12},
  pages={2295--2329},
  year={2017},
  publisher={Ieee}
}

@article{sharma2017activation,
  title={Activation functions in neural networks},
  author={Sharma, Sagar and Sharma, Simone and Athaiya, Anidhya},
  journal={Towards Data Sci},
  volume={6},
  number={12},
  pages={310--316},
  year={2017}
}

@inproceedings{boureau2010theoretical,
  title={A theoretical analysis of feature pooling in visual recognition},
  author={Boureau, Y-Lan and Ponce, Jean and LeCun, Yann},
  booktitle={Proceedings of the 27th international conference on machine learning (ICML-10)},
  pages={111--118},
  year={2010}
}

@article{hochreiter1997long,
  title={Long short-term memory},
  author={Hochreiter, Sepp and Schmidhuber, J{\"u}rgen},
  journal={Neural computation},
  volume={9},
  number={8},
  pages={1735--1780},
  year={1997},
  publisher={MIT press}
}

@inproceedings{graves2005framewise,
  author    = {Graves, Alex and Schl{\"u}ter, Nicol and Fern{\'a}ndez, Santiago and Schmidhuber, J{\"u}rgen},
  title     = {Framewise Phoneme Classification with Bidirectional {LSTM} and Other Neural Network Architectures},
  booktitle = {Proceedings of the IEEE International Joint Conference on Neural Networks (IJCNN)},
  year      = {2005},
  pages     = {2047--2052},
  doi       = {10.1109/IJCNN.2005.1556215}
}

@article{srivastava2014dropout,
  title={Dropout: a simple way to prevent neural networks from overfitting},
  author={Srivastava, Nitish and Hinton, Geoffrey and Krizhevsky, Alex and Sutskever, Ilya and Salakhutdinov, Ruslan},
  journal={The journal of machine learning research},
  volume={15},
  number={1},
  pages={1929--1958},
  year={2014},
  publisher={JMLR. org}
}

@inproceedings{bridle1990probabilistic,
  author    = {Bridle, John S.},
  title     = {Probabilistic Interpretation of Feedforward Classification Network Outputs, with Relationships to Statistical Pattern Recognition},
  booktitle = {Neurocomputing: Algorithms, Architectures and Applications},
  year      = {1990},
  pages     = {227--236}
}

@article{kingma2014adam,
  title={Adam: A method for stochastic optimization},
  author={Kingma, Diederik P and Ba, Jimmy},
  journal={arXiv preprint arXiv:1412.6980},
  year={2014}
}

@article{cuesta2010use,
  title={{The use of inertial sensors system for human motion analysis}},
  author={Cuesta-Vargas, Antonio I and Gal{\'a}n-Mercant, Alejandro and Williams, Jonathan M},
  journal={Physical Therapy Reviews},
  volume={15},
  number={6},
  pages={462--473},
  year={2010},
  publisher={Taylor \& Francis}
}

@article{camomilla2018trends,
  title={{Trends supporting the in-field use of wearable inertial sensors for sport performance evaluation: A systematic review}},
  author={Camomilla, Valentina and Bergamini, Elena and Fantozzi, Silvia and Vannozzi, Giuseppe},
  journal={Sensors},
  volume={18},
  number={3},
  pages={873},
  year={2018},
  publisher={MDPI}
}

@article{klein2019smartphone,
  title={Smartphone location recognition: A deep learning-based approach},
  author={Klein, Itzik},
  journal={Sensors},
  volume={20},
  number={1},
  pages={214},
  year={2019},
  publisher={MDPI}
}

@article{gu2021survey,
  title={A survey on deep learning for human activity recognition},
  author={Gu, Fuqiang and Chung, Mu-Huan and Chignell, Mark and Valaee, Shahrokh and Zhou, Baoding and Liu, Xue},
  journal={ACM Computing Surveys (CSUR)},
  volume={54},
  number={8},
  pages={1--34},
  year={2021},
  publisher={ACM New York, NY}
}

@article{kaseris2024survey,
  title={A Comprehensive Survey on Deep Learning Methods in Human Activity Recognition},
  author={Kaseris, Markos and Kampianakis, Emmanouil and Tzes, Anthony},
  journal={Machine Learning and Knowledge Extraction},
  volume={6},
  number={2},
  pages={582--615},
  year={2024},
  publisher={MDPI}
}

@article{gomaa2023perspective,
  title={A perspective on human activity recognition from inertial motion data},
  author={Gomaa, Ahmed and Khamis, Aly},
  journal={IEEE Sensors Journal},
  volume={23},
  number={21},
  pages={24377--24395},
  year={2023},
  publisher={IEEE}
}

@article{abumostafa1989information,
  title={Information theory, complexity, and neural networks},
  author={Abu-Mostafa, Yaser S},
  journal={IEEE Communications Magazine},
  volume={27},
  number={11},
  pages={25--28},
  year={1989},
  publisher={IEEE},
  note={Origin of the 'Rule of 10' heuristic}
}

@article{raudys1991small,
  title={Small sample size effects in statistical pattern recognition: Recommendations for practitioners},
  author={Raudys, Sarunas J and Jain, Anil K},
  journal={IEEE Transactions on Pattern Analysis and Machine Intelligence},
  volume={13},
  number={3},
  pages={252--264},
  year={1991},
  publisher={IEEE}
}

@article{balki2019sample,
  title={Sample-size determination methodologies for machine learning in medical imaging research: a systematic review},
  author={Balki, Inderbir and Amirabadi, Afsaneh and Levman, Jacob and Martel, Anne L and Emersic, Ziga and Meden, Blaz and Garcia-Pedrero, Angel and Ramirez, S C and Kong, D and Moody, A R and Tyrrell, P N},
  journal={Canadian Association of Radiologists Journal},
  volume={70},
  number={4},
  pages={344--353},
  year={2019},
  publisher={Elsevier},
  doi={10.1016/j.carj.2019.06.002},
  note={Key review identifying the lack of rigorous sample size criteria in ML studies}
}

@article{weiss2019wisdm,
  title={Wisdm smartphone and smartwatch activity and biometrics dataset},
  author={Weiss, Gary M},
  journal={UCI Machine Learning Repository: WISDM Smartphone and Smartwatch Activity and Biometrics Dataset Data Set},
  volume={7},
  number={133190-133202},
  pages={5},
  year={2019},
  publisher={Springer}
}

@inproceedings{anguita2013public,
  title={A public domain dataset for human activity recognition using smartphones.},
  author={Anguita, Davide and Ghio, Alessandro and Oneto, Luca and Parra, Xavier and Reyes-Ortiz, Jorge Luis and others},
  booktitle={Esann},
  volume={3},
  number={1},
  pages={3--4},
  year={2013}
}

@article{cohen2024inertial,
  title={Inertial navigation meets deep learning: A survey of current trends and future directions},
  author={Cohen, Nadav and Klein, Itzik},
  journal={Results in Engineering},
  pages={103565},
  year={2024},
  publisher={Elsevier}
}

@inproceedings{yampolsky2025neural,
  title={On Neural Inertial Classification Networks for Pedestrian Activity Recognition},
  author={Yampolsky, Zeev and Kruzel, Ofir and Fekson, Victoria Khalfin and Klein, Itzik},
  booktitle={2025 IEEE/ION Position, Location and Navigation Symposium (PLANS)},
  pages={15--22},
  year={2025},
  organization={IEEE}
}

@article{fekson2026enhancement,
  title={Enhancement of neural inertial regression networks: A data-driven perspective},
  author={Fekson, Victoria Khalfin and Pri-Hadash, Nitsan and Palez, Netta and Etzion, Aviad and Klein, Itzik},
  journal={Results in Engineering},
  pages={109727},
  year={2026},
  publisher={Elsevier}
}

@article{kruzel2026optimizing,
  title={Optimizing Neural Inertial Classification: A Benchmark Study of Data-Driven Techniques},
  author={Kruzel, Ofir and Klein, Itzik and others},
  journal={IEEE Journal of Indoor and Seamless Positioning and Navigation},
  year={2026},
  publisher={IEEE}
}

@article{althnian2021impact,
  title={Impact of dataset size on classification performance: an empirical evaluation in the medical domain},
  author={Althnian, Alhanoof and AlSaeed, Duaa and Al-Baity, Heyam and Samha, Amani and Dris, Alanoud Bin and Alzakari, Najla and Abou Elwafa, Afnan and Kurdi, Heba},
  journal={Applied sciences},
  volume={11},
  number={2},
  pages={796},
  year={2021},
  publisher={MDPI}
}

@article{barbedo2018impact,
  title={Impact of dataset size and variety on the effectiveness of deep learning and transfer learning for plant disease classification},
  author={Barbedo, Jayme Garcia Arnal},
  journal={Computers and electronics in agriculture},
  volume={153},
  pages={46--53},
  year={2018},
  publisher={Elsevier}
}

@article{klein2025pedestrian,
  title={Pedestrian inertial navigation: An overview of model and data-driven approaches},
  author={Klein, Itzik},
  journal={Results in Engineering},
  volume={25},
  pages={104077},
  year={2025},
  publisher={Elsevier}
}

@article{vertzberger2021attitude,
  title={Attitude adaptive estimation with smartphone classification for pedestrian navigation},
  author={Vertzberger, Eran and Klein, Itzik},
  journal={IEEE Sensors Journal},
  volume={21},
  number={7},
  pages={9341--9348},
  year={2021},
  publisher={IEEE}
}

@article{MARDANPOUR2023119073,
title = {Human activity recognition based on multiple inertial sensors through feature-based knowledge distillation paradigm},
journal = {Information Sciences},
volume = {640},
pages = {119073},
year = {2023},
issn = {0020-0255},
doi = {https://doi.org/10.1016/j.ins.2023.119073},
url = {https://www.sciencedirect.com/science/article/pii/S0020025523006588},
author = {Malihe Mardanpour and Majid Sepahvand and Fardin Abdali-Mohammadi and Mahya Nikouei and Homeyra Sarabi}
}

@Article{make6020040,
AUTHOR = {Kaseris, Michail and Kostavelis, Ioannis and Malassiotis, Sotiris},
TITLE = {A Comprehensive Survey on Deep Learning Methods in Human Activity Recognition},
JOURNAL = {Machine Learning and Knowledge Extraction},
VOLUME = {6},
YEAR = {2024},
NUMBER = {2},
PAGES = {842--876},
URL = {https://www.mdpi.com/2504-4990/6/2/40},
ISSN = {2504-4990},
DOI = {10.3390/make6020040}
}

@article{abbas2026bibliometric,
  author  = {Abbas, I. and et al.},
  title   = {A bibliometric review of deep learning in crop monitoring: trends, challenges, and future perspectives},
  journal = {PMC},
  year    = {2026},
  volume  = {12484012},
  doi     = {10.3390/app11020796}
}

\end{document}